\documentclass[11pt]{article}
\usepackage{acl2014}
\usepackage{times}
\usepackage{url}
\usepackage{latexsym}

\usepackage{graphicx}

\usepackage{tipa}

\usepackage{multirow}
\usepackage{booktabs}
\usepackage{amsmath}
\usepackage{amsfonts}

\newcommand{\dtw}{\textrm{DTW}}
\newcommand{\dba}{\textrm{DBA}}

\let\citet\newcite
\let\citep\cite

\title{A case study on using speech-to-translation alignments for language documentation}

\author{Antonios Anastasopoulos \\
 University of Notre Dame \\
  {\tt aanastas@nd.edu} \\
  \And
  David Chiang \\
  University of Notre Dame\\
  {\tt dchiang@nd.edu} \\}

\date{}

\begin{document}
\maketitle
\begin{abstract}
  For many low-resource or endangered languages, spoken language resources are more likely to be annotated with translations than with transcriptions. Recent work exploits such annotations to produce speech-to-translation alignments, without access to any text transcriptions.
  We investigate whether providing such information can aid in producing better (mismatched) crowdsourced transcriptions, which in turn could be valuable for training speech recognition systems, and show that they can indeed be beneficial through a small-scale case study as a proof-of-concept.
  We also present a simple phonetically aware string averaging technique that produces transcriptions of higher quality.
\end{abstract}

\section{Introduction}

For many low-resource and endangered languages, speech data is easier to obtain than textual data. 
The traditional method for documenting a language involves a trained linguist collecting speech and then transcribing it, often at a phonetic level, as most of these languages do not have a writing system. This, however, is a costly and slow process, as it could take up to 1 hour for a trained linguist to transcribe the phonemes of 1 minute of speech \cite{thitowards}.

Therefore, speech is more likely to be annotated with translations than with transcriptions. This translated speech is a potentially valuable source of information as it will make the collected corpus interpretable for future studies. New technologies are being developed to facilitate collection of translations \cite{bird+al:2014}, and there already exist recent examples of parallel speech collection efforts focused on endangered languages \cite{blachon2016parallel,adda2016breaking}. 

Recent work relies on parallel speech in order to create speech-to-translation alignments \cite{anastasopoulos-chiang-duong:2016:EMNLP2016}, discover spoken terms \cite{bansal2016weakly,godard2016preliminary}, learn a lexicon and translation model \cite{adams-EtAl:2016:EMNLP2016}, or directly translate speech \cite{long-EtAl:2016:NAACL-HLT,berard2016listen}. 
Another line of work \cite{das2016automatic,jyothi2015transcribing,liu2016adapting} focuses on training speech recognition systems for low-resource settings using mismatched crowdsoursed transcriptions. These are transcriptions that include some level of noise, as they are crowdsourced from workers unfamiliar with the language being spoken.

We aim to explore whether the quality of crowdsourced transcriptions could benefit from providing transcribers with speech-to-translation word-level alignments. That way, speech recognition systems trained on the higher-quality probabilistic transcriptions (of at least a sample of the collected data) could be used as part of the pipeline to document an endangered language.

\section{Methodology}

As a proof-of-concept, we work on the language pair Griko-Italian, for which there exists a sentence-aligned parallel corpus of source-language speech and target-language text \cite{grikodatabase}. Griko is an endangered minority language spoken in the south of Italy. Using the method of \citet{anastasopoulos-chiang-duong:2016:EMNLP2016}, we also obtain speech-to-translation word-level alignments.

The corpus that we work on already provides gold-standard transcriptions and speech-to-translation alignments, so it is suitable for conducting a case study that will examine the potential effect of providing the alignments on the crowdsourced transcriptions, as we will be able to compare directly against the gold standard.

We randomly sampled~30 utterances from the corpus and collected transcriptions through a simple online interface (described at \S\ref{sec:interface}) from 12 different participants. None of the participants spoke or had any familiarity with Griko or its directly related language, Greek. Six of the participants were native speakers of Italian, the language in which the translations are provided.  Three of them did not speak Italian, but were native Spanish speakers, and the last 3 were native English speakers who also did not speak Italian but had some level of familiarity with Spanish.

The~30 utterances amount to~$1.5$ minutes of speech, which would potentially require~$1.5$ hours of a trained linguist's work to phonetically transcribe. The gold Griko transcriptions include~191 Griko tokens, with 108 types. Their average length is~$6.5$ words, with the shortest being~2 words and the longest being~14 words.

The utterances were presented to the participants in three different modes:
\begin{enumerate}
    \item \texttt{no} mode: Only providing the translation text.
    \item \texttt{auto} mode: Providing the translation text and the potentially noisy speech-to-translation alignments produced by the method of \citet{anastasopoulos-chiang-duong:2016:EMNLP2016}.
    \item \texttt{gold} mode: Providing the translation text and the gold-standard speech-to-translation alignments.
\end{enumerate}

The utterances were presented to the participants in the exact same order, but in different modes following a  scheme according to the utterance id (1 to~30) and the participant id (1 to~12). The first utterance was transcribed by the first participant under \texttt{no} mode, by the second participant under \texttt{auto} mode, the third participant under \texttt{gold} mode, the fourth participant under \texttt{no} mode, etc. The second utterance was presented to the first participant under \texttt{auto} mode, to the second participant under \texttt{gold} mode, to the third participant under \texttt{no} mode, etc. 

This rotation scheme ensured that the utterances were effectively split into~3 subsets, each of which was transcribed exactly~4 times in each mode, with~2 of them by an Italian speaker,~1 time by a Spanish speaker, and~1 time by an English speaker. This enables a direct comparison of the three modes, and, hopefully, an explanation of the effect of providing the alignments. The modes under which each participant had to transcribe the utterances changed from one utterance to another, in order to minimize the potential effect of the participants' learning of the task and the language better.

The participants were asked to produce a transcription of the given speech segment, using the Latin alphabet and any pronunciation conventions they wanted. The result in almost all cases is entirely comprised of nonsense syllables. It is safe to assume, though, that the participants would use the pronunciation conventions of their native language; for example, an Italian or Spanish speaker would transcribe the sounds \textipa{[mu]} as \texttt{mu}, whereas an English native speaker would probably transcribe it as \texttt{moo}.

\section{Interface}
\label{sec:interface}

\begin{figure*}
    \centering
    \includegraphics[scale=0.3]{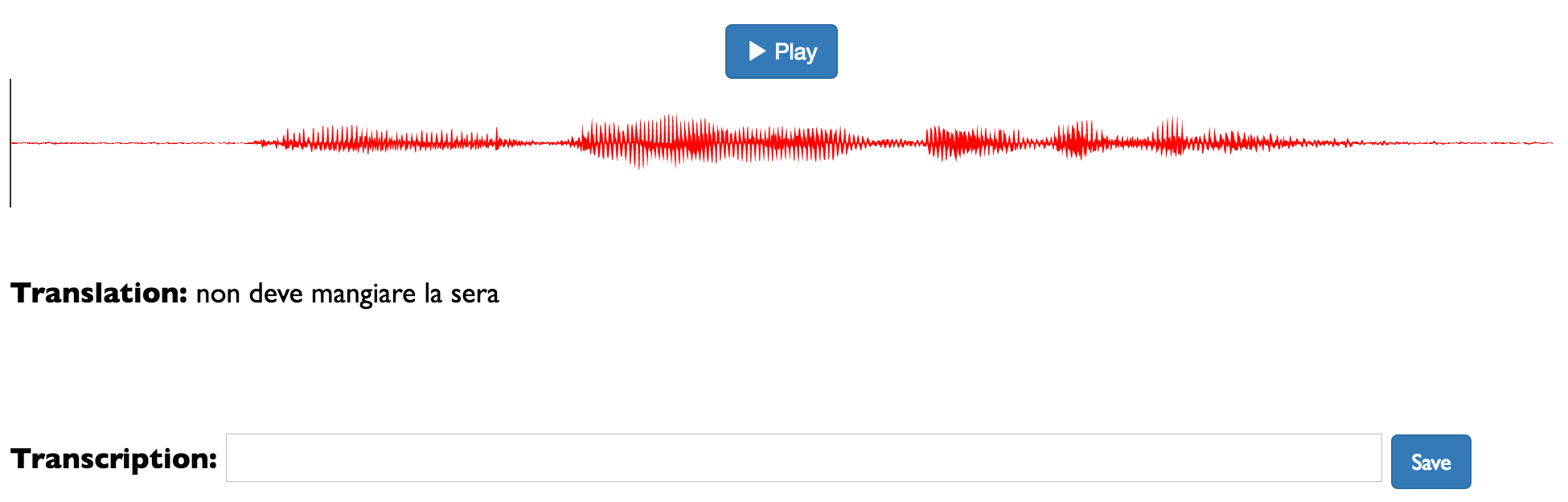}
    \caption{Interface that only provides the translation \texttt{non deve mangiare la sera} \textit{[he/she shouldn't eat at night]}, with no alignment information.}
    \label{fig:simple_interface}
    \includegraphics[scale=0.3]{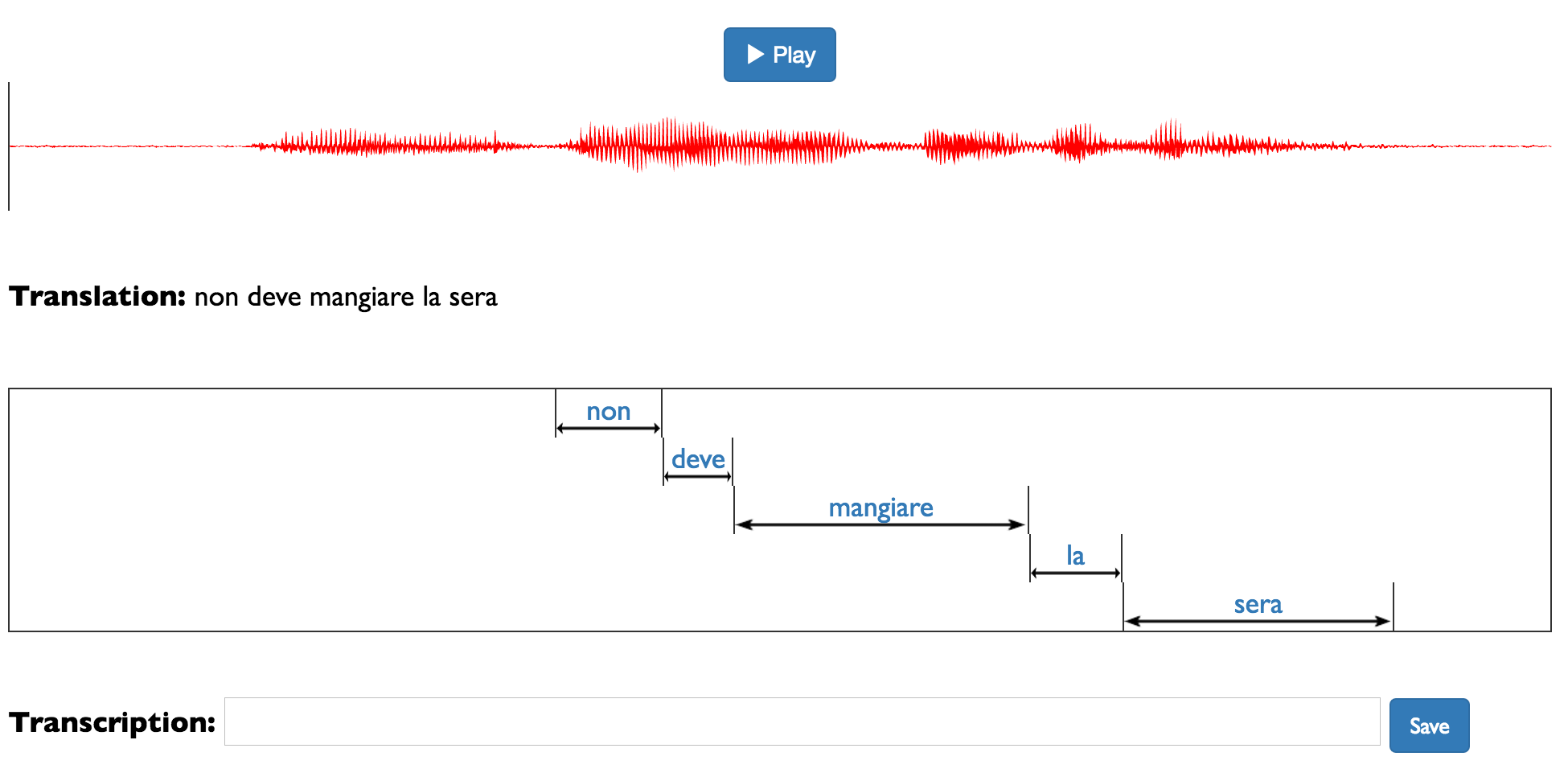}
    \caption{Interface that provides the translation \texttt{non deve mangiare la sera} \textit{[he/she shouldn't eat at night]}, along with speech-to-translation alignment information. Clicking on a translation word would play the corresponding aligned part of the speech segment.}
    \label{fig:al_interface}
\end{figure*}

A simple tool for collecting transcriptions first needs to provide the user with the audio to be transcribed. The translation of the spoken utterance is provided, as in Figure~\ref{fig:simple_interface}, where in our case the speech to be transcribed was in Griko, and a translation of this segment was provided in Italian. In a real scenario, this translation would correspond to the output of a Speech Recognition system for the parallel speech, so it could potentially be somewhat noisy. Though, for the purposes of our case study, we used the gold standard translations of the utterances.

Our interface also provides speech-to-translation alignment information as shown in Figure~\ref{fig:al_interface}. Each word in the translation has been aligned to some part of the spoken utterance. Apart from listening to the whole utterance at once, the user can also click on the individual translation words and listen to the corresponding speech segment.

For the purposes of our case study, our tool collected additional information about its usage. It logged the amount of time each participant spent transcribing each utterance, as well as the amount of times that they clicked the respective buttons in order to listen to either the whole utterance or word-aligned speech segments.

\section{Results}

The orthography of Griko is phonetic, and therefore it is easy, using simple rules, to produce the phonetic sequences in IPA that correspond to the transcriptions. We can also use standard rules for Spanish (LDC96S35) and Italian,\footnote{Creating the rules based on \cite{comrie2009world}} depending on the native language of the participants, in order to produce phonetic sequences of the crowdsourced transcriptions in IPA. 

For simplicity reasons, we merge the vowel oppositions /\textipa{e}$\sim$\textipa{E}/ and /\textipa{o}$\sim$\textipa{O}/ into just /\textipa{e}/ and /\textipa{o}/ for both the Italian and Griko phonetic transcriptions, as neither of the two languages makes an orthographic distinction between the two.

For the transcriptions created by the English-speaking participants, and since most of the word-like units of the transcriptions do not exist in any English pronunciations lexicon, we use the LOGIOS Lexicon Tool \cite{logios} that uses some simple letter-to-sound rules to produce a phonetic transcription in the ARPAbet symbol set. We map several of the English vowel oppositions to a single IPA vowel; for example, \texttt{IH} and \texttt{IY} both become \textipa{/i/}, while \texttt{UH} and \texttt{UW} become \textipa{/u/}. Phonemes \texttt{AY}, \texttt{EY}, and \texttt{OY} become \textipa{/ai/}, \textipa{/ei/}, and \textipa{/oi/} respectively. This enables a direct comparison of all the transcriptions, although it might add extra noise, especially in the case of transcriptions produced by English-speaking participants.

Two examples of the resulting phonetic transcriptions as produced by the participants' transcriptions can be found in Tables~\ref{tab:example1} and \ref{tab:example2}. 

\begin{table}[h]
    \centering
    \begin{tabular}{c|c|c|c}
        \toprule
         & transcription  & distance \\
         \midrule
        \texttt{it1} & \textipa{o ladro isodzeem biabiddu} & 5  \\
        \texttt{it2} & \textipa{o ladro isodZenti dabol tu} & 6 \\
        \texttt{it3}& \textipa{o ladro i so ndze mia buttu} & 5\\
        \texttt{it4}& \textipa{o ladro isodzeembia po tu}& 2\\
        \texttt{it5}& \textipa{o ladroi isodZe enbi a buttu} & 4\\
        \texttt{it6}& \textipa{o ladro idZo dzembia a buttu} & 7\\
        \texttt{es1}& \textipa{o la vro ipsa ziem biabotu} &  9\\
        \texttt{es2}& \textipa{ola avro isonse embia butu} &  7\\
        \texttt{es3}& \textipa{o ladro isosen be abuto} & 9\\
        \texttt{en1}& \textipa{o labro ebzozaim bellato} & 13 \\
        \texttt{en2}& \textipa{o laha dro iso dzenne da to} & 12 \\
        \texttt{en3}& \textipa{o ladro i dzo ze en habito} & 11\\
        \midrule
        average & \textipa{o ladro isodZe mbia buttu} & 3\\
        \midrule
        correct & \textipa{o ladro isodZe embi apo ttu} & \\
        \bottomrule
    \end{tabular}
    \caption{Transcriptions for the utterance \texttt{o l\`adro \`isoze \`embi apo-tt\`u} \textit{[the thief must have entered from here]} and their Levenshtein distance to the gold transcription. The word \texttt{ladro} \textit{[thief]} is the same in both Griko and Italian.}
    \label{tab:example1}
\end{table}

\begin{table*}
\centering
    \begin{tabular}{c|c|c}
        \toprule
        participant & acoustic transcription  & distance \\
         \midrule
        \texttt{it1} & \textipa{bau tSerkianta ena furno e tranni e rustiku} & 9 \\
        \texttt{it2} & \textipa{pau tSerkianta ena furna kanni e rustiku} & 7\\
        \texttt{it3} & \textipa{pau tSerkianta na furno kakanni rustiko} & 5\\
        \texttt{it4} & \textipa{po Serkieunta na furna ka kanni rustiku} & 6\\
        \texttt{it5}& \textipa{pau tSerkeunta en furno ganni rustiku} & 6\\
        \texttt{it6}& \textipa{pa u tSerkionta en na furno kahanni rustiko} & 5\\
        \texttt{es1}& \textipa{pogurSe kiunta en a furna e kakani e rustiku} &  12\\
        \texttt{es2}& \textipa{pao Serkeonta ena furna ka kani rustigo} &  5\\
        \texttt{es3} & \textipa{bao tSerke on ta e na furno e kagani e rustiko} & 6\\
        \texttt{en1} & \textipa{paoje kallonta e un forno e grane e rustiko} & 15\\
        \texttt{en2} & \textipa{pao tSerkeota eno furno e kakarni e rustiko} & 5\\
        \texttt{en3} & \textipa{pouSa kianta e a forno e tagani e rustiko} & 14\\
        \midrule
        average & \textipa{pao tSerkionta ena furno kaanni e rustiku} & 3\\
        \midrule
        correct & \textipa{pao tSerkeonta ena furno ka kanni rustiku} & \\
        \bottomrule
    \end{tabular}
    \caption{Transcriptions for the utterance \texttt{p\`ao cerk\`eonta \`ena f\`urno ka k\`anni r\`ustiku} \textit{[I'm looking for a bakery that makes rustic (bread)]} and their Levenshtein distance to the gold transcription.}
    \label{tab:example2}
\end{table*}

On our analysis of the results, we first focus on the results obtained by the 6 Italian-speaking participants of our study, which represent the more realistic crowdsourcing scenario where the workers speak the language of the translations. We then present the results of the non-Italian speaking participants. In order to evaluate the transcriptions, we report the Levenshtein distance as well as the average Phone Error Rate (PER)\footnote{The Phone Error Rate is basically length-normalized Levenshtein distance.} against the correct transcriptions.

\subsection{Italian-speaking participants}

\begin{table}
    \centering
    \begin{tabular}{c|ccc|c}
    \toprule
    utterance & \multicolumn{4}{c}{Levenshtein distance} \\
    set & \texttt{no}  & \texttt{auto}  & \texttt{gold}  & all modes\\
    \midrule
    set 1 & 14.1 & 13.5 & 13.9 & 13.8\\
    set 2 & 10.0 & 10.6 & 8.7 & 9.8\\
    set 3 & 11.8 & 10.1 & 10.5 & 10.8\\
    \midrule
    average & 12.0 & 11.4 & 11.0 & 11.5 \\
    \bottomrule
    \end{tabular}
    \caption{Breakdown of the quality of the transcriptions per utterance set. The value in each cell corresponds to the average Levenshtein distance to the gold transcriptions. Despite the differences in how ``hard" each set is, the transcription quality generally improves when alignments are provided, as shown by the average in the last row.}
    \label{tab:levenshtein-it}
    \end{table}
\begin{table}
    \centering
    \begin{tabular}{c|ccc|c}
    \toprule
    utterance & \multicolumn{4}{c}{PER } \\
    set & \texttt{no} & \texttt{auto} & \texttt{gold} & all modes\\
    \midrule
    set 1 & 23.0 & 25.1 & 23.8 & 24.0 \\
    set 2 & 25.8 & 26.0 & 23.3 & 25.0\\
    set 3 & 32.1 & 26.0 & 24.5 & 28.1\\
    \midrule
    avg  & 27.0 & 25.7 & 24.5 & 25.7 \\
    \bottomrule
    \end{tabular}
    \caption{Phone Error Rate (PER) of the phonetic transcriptions produced by the Italian-speaking participants per utterance set. In the general case, the quality improves when alignments are provided, as shown by the averages in the last row. }
    \label{tab:per-it}
\end{table}

\paragraph{Transcription quality} 

As a first test, we compare the Levenshtein distances of the produced transcriptions to the gold ones. For fairness, we remove the accents from the gold Griko transcriptions, as well as any accents added by the Italian speaking participants. 

The results averaged per utterance set and per mode are shown in Table~\ref{tab:levenshtein-it}. We first note that the three utterance sets are not equally hard: the first one is the hardest, with the second one being the easiest one to transcribe, as it included slightly shorter sentences. However, in most cases, as well as in the average case (last row of Table~\ref{tab:levenshtein-it}) providing the alignments improves the transcription quality. In addition, the gold standard alignments provide more accurate information that is also reflected in higher quality transcriptions. 

We also evaluate the precision and recall of the word boundaries (spaces) that the transcriptions denote. We count a discovered word boundary as a correct one only if the word boundary in the transcription is matched with a boundary marker in the gold transcription, when we compute the Levenshtein distance. 

Under \texttt{no} mode (without alignments), the transcribers achieve~58\% recall and~70\% precision on correct word boundaries. However, when provided with alignments, they achieve~66\% recall and~77\% precision; in fact, when provided with gold alignments (under \texttt{gold} mode) recall increases to~70\% and precision to~81\%. Therefore, the speech-to-translation alignments seem to provide 
information that helped the transcribers to better identify word boundaries, which is arguably hard to achieve from just continuous speech.

\paragraph{Phonetic transcription quality} We observe the same pattern when evaluating 
using the average PER of these phonetic sequences, as reported in Table~\ref{tab:per-it}:  the acoustic transcriptions 
%(either in their original orthography, or converted to phonetic ones in IPA) 
are generally better when alignments are provided. Also, the gold alignments provide more accurate information, resulting in higher quality transcriptions. However, even using the noisy alignments leads to better transcriptions in most cases.

It is worth noting that out of the~30 utterances, only~4 included words that are shared between Italian and Griko (\texttt{ancora}~\textit{[yet]}, \texttt{ladro}~\textit{[thief]}, \texttt{giornale}~\textit{[newspaper]}, and \texttt{subito}~\textit{[immediately]} ) and only~2 of them included common proper names (\texttt{Valeria} and \texttt{Anna}). The effect of having those common words, therefore, is minimal.

\subsection{Non-Italian speaking participants}

The scenario where the crowdsourcers do not even speak the language of the translations is possibly too extreme. It still could be applicable, though, in the case where the language of the translations is not endangered by still low-resource (Tok Pisin, for example) and it's hard to find annotators that speak the language. In any case, we show that if the participants speak a language related to the translations (and with a similar phonetic inventory, like Spanish in our case) they can still produce decent transcriptions.

\begin{table}[ht]
    \centering
    \begin{tabular}{c|c}
        \toprule
        participants & PER  \\
        \midrule
        Italian & 25.7\\
        Spanish & 28.3\\
        English & 34.3\\
        \midrule
        all & 28.5 \\
        \midrule
        best & 22.8\\
        worst & 37.0\\
        \bottomrule
    \end{tabular}
    \caption{Breakdown of the quality of the transcriptions per participant group. As expected, the group of participants that speak the language closest to the target language (Italian) produces better transcriptions.}
    \label{tab:nationality}
\end{table}

Table~\ref{tab:nationality} shows the average on the performance of the different groups of participants. As expected, the Italian-speaking participants produced higher quality transcriptions, but the Spanish-speaking participants did not perform much worse. Also in the case of non-Italian speaking participants, we found that providing speech-to-translation alignments (under \texttt{auto} and \texttt{gold} modes) improves the quality of the transcriptions, as we observed a similar trend as the ones shown in Tables~\ref{tab:levenshtein-it} and~\ref{tab:per-it}. 

The noise in the non-Italian speaker annotations, and especially the ones produced by English speakers, can be explained in two ways. One, it could be caused by annotation scheme employed by the English speakers, which must be more complicated and noisy, as English does not have a concrete letter-to-sound system.  Or two, it could be explained by the fact that English is much more typologically distant from Griko, meaning, possibly, that some of the sounds in Griko just weren't accessible to English speakers. The latter effect could indeed be real, as it has been shown that a language's phonotactics can affect what sounds a speaker is actually able to perceive ~\cite{peperkamp1999perception,dupoux2008persistent}.  The perceptual ``illusions" created by one's language can be quite difficult to overcome.

\subsection{Overall discussion}

From the results, it is clear that the acoustic transcriptions 
%(either in their original orthography, or converted to phonetic ones in IPA) 
are generally better when collected with the alignments provided. Also, the gold alignments provide more accurate information, resulting in higher quality transcriptions. However, even using the noisy alignments leads to better transcriptions in most cases.

One simple explanation for this finding is that our interface changes when we provide alignments, giving the participants an easier way to listen to much shorter segments of the speech utterance. Therefore, our observations of improved transcriptions might not be caused because of the alignments, but because of the change in the interface. This can be tested by comparing results obtained by two interfaces, one that is similar to ours, providing the alignments, and one that also provides the option to play shorter segments of the speech utterance, randomly selected. We leave however this test to be performed in a future study.

The results in Tables~\ref{tab:levenshtein-it} and \ref{tab:per-it}
are indicative of how, on average, we can collect better transcriptions by providing speech-to-translation alignments. However, we could obtain a better understanding by comparing the transcription modes on each individual utterance level. 
%but the individual users' results are not directly comparable because the sets of utterances that each user transcribed in each mode are not overlapping. 
%but we could obtain a better understanding by comparing the transcription modes on each individual utterance level. 

For each utterance we have in total~12 transcriptions,~4 for each mode. We therefore have~48 possible combinations of pairs of transcriptions of the same utterance that were performed under a different mode. 
This means that we can have $48 \times 30 = 1440$ pairwise comparisons in total (so that the pairs include only transcriptions of the same utterance). 
In the overwhelming majority of these comparisons (73\%) the transcriptions obtained with alignments provided, were better than the ones obtained without them. 

In addition, for the about~380 pairs where the transcription obtained without alignments is better than the one obtained with alignments, the majority corresponds to pairs that include a combination of an Italian speaking participant (without alignments) and a Spanish or English speaking participant (with alignments). 
For example, the very meticulous participant \texttt{it4} (who in fact achieves the shortest distance to the gold transcriptions) provides in several cases better transcriptions than almost all English and Spanish speaking participants, even without access to speech-to-translation alignments.

\paragraph{Time} It took about~36 minutes on average for the 12 participants to complete the study (shortest was 20 minutes, longest was 64 minutes). This is less time than what trained linguists typically require \cite{thitowards}, at the expense, naturally, of much higher error rates.
%This is significantly less time than what a trained linguist would probably need, and at much lower potential cost, at the expense, naturally, of a few errors in the transcriptions.

At an utterance level, we find that providing the participants with the alignment information does not impact the time required to create the transcription. When provided with alignments, the participants listened to the whole utterance about 30\% fewer times; instead, they chose to click on and play alignment segments almost as many times as opting to listen to the whole utterance. There was only one participant who rarely chose to play the alignment segment, and in fact the average quality of their transcriptions does not differ across the different modes. 

\section{Averaging the acoustic transcriptions}

A fairly simple way to merge several transcriptions into one, is to obtain first alignments between the set of strings to be averaged by treating each substitution, insertion, deletion, or match, as an alignment. Then, we can leverage the alignments in order to create an ``average" string, through an averaging scheme.

We propose a method that can be roughly described as similar to using Dynamic Time Warping (\dtw{}) \cite{berndt1994using} for obtaining alignments between two speech signals, and using \dtw{} Barycenter Averaging (DBA) \cite{petitjean2011global} for approximating the average of a set of sequences. Instead of time series or speech utterances, however, we apply these methods on sequences of phone embeddings.

We map each IPA phone into a feature embedding, with boolean features corresponding to linguistic features.\footnote{The features were taken from the inventories of \url{http://phoible.org/}} Then, each acoustic transcription can be represented as a sequence of vectors, and we can use \dba{} in order to obtain an ``average" sequence, out of a set of sequences. This ``average" sequence can be then mapped back to phones, by mapping each vector to the phone that has the closest phone embedding in our space.

The standard method, ROVER~\cite{fiscus1997post}, uses an alignment module and then majority voting to produce a probabilistic final transcription. The string averaging method that we propose here is quite similar, with the exception that our alignment method and the averaging method are tied together through the iterative procedure of \dba{}. Another difference is that our method operates on phone embeddings, instead of directly on phones. That way, it is more phonologically informed, so that the distance between two phones that are often confused because they have similar characteristics, such as \textipa{/p/} and \textipa{/b/}, is smaller than the distance between a pair of more distant phones such as eg. \textipa{/p/} and \textipa{/a/}. In addition, the averaging scheme that we employ actually produces an average of the aligned phone embeddings, which in theory could result in a different output compared to simple majority voting. A more thorough comparison of ROVER and our averaging method is beyond the scope of this paper and is left as future work.

Using this simple string averaging method  we combine the mismatched transcriptions into an ``average" one. We can then compute the Levenshtein distance and PER between the ``average" and the gold transcription in order to evaluate them. Examples of ``average" transcriptions are also shown in Tables~\ref{tab:example1} and \ref{tab:example2}. In almost all cases the ``average" transcription is closer to the gold one than each of the individual transcriptions. Table~\ref{tab:averages} provides a more detailed analysis of the quality of the ``average" transcriptions per mode and per group of participants. 

We first use the transcriptions as produced by all participants, and report the errors of the averaged outputs under all modes. Again, the transcriptions that were produced with alignments provided, when averaged, have lower error rates. However, the \texttt{gold} mode corresponds to an ideal scenario, which will hardly ever occur. Thus, we focus more on the combination of the \texttt{no} and \texttt{auto} modes, which will very likely occur in our collection efforts, as the alignments we will produce will be noisy, or we might only have translations without alignments. We also limit the input to only include the transcriptions produced by the Italian and Spanish speaking participants, as we found that the transcriptions produced by English speaking participants added more noise instead of helping. As the results in Table~\ref{tab:averages} show, using our averaging method we obtain better transcriptions on average, even if we limit ourselves to the more realistic scenario of not having \texttt{gold} alignments. The best result with an average PER of~$23.2$ is achieved using all the transcriptions produced by Italian and Spanish speaking participants. Even without using gold alignments, however, the averaging method produces transcriptions that achieve an average PER of 24.0, which is a clear improvement over the average PER of the individual transcriptions ($25.7$).

The reason that the ``average" transcription is better than the transcriptions used to create it is intuitive. Although all the  transcriptions include some level of noise, not all of the transcribers make the same errors. Averaging the produced transcriptions together helps overcome most of the errors, simply because the majority of the participants does not make each individual error. This is, besides, the intuition behind the previous work on using mismatched crowdsourced transcriptions. 
In addition, one of the bases of the Aikuma approach \citep{Bird:2010:SMP:1875689.1875692} to language documentation is re-speaking of the original text. Our averaging method could potentially also be applied to transcriptions obtained from these re-spoken utterances, further improving the quality of the transcriptions.

\begin{table}
    \centering
    \begin{tabular}{c|c|c|c}
        \toprule
        \multicolumn{2}{c|}{transcriptions used }& \multicolumn{2}{|c}{avg. distance}\\
        \multicolumn{2}{c|}{to create average} & \multicolumn{2}{|c}{to gold}\\ 
        \multirow{2}{*}{mode} & participants' & \multirow{2}{*}{Lev/tein} & \multirow{2}{*}{PER}\\
        & native language & &\\
        \midrule
        \texttt{no} & all & 8.41 & 27.0\\
        \texttt{auto} & all & 7.82 & 25.9\\
        \texttt{gold} & all & 7.58 & 24.3\\
        \midrule
        all & Ita+Spa & \textbf{7.21} & \textbf{23.2} \\
        \texttt{gold} & Ita+Spa & 7.55 & 23.6 \\
        \texttt{no+auto} & Ita+Spa & 7.62 & 24.0 \\
        \bottomrule
    \end{tabular}
    \caption{Average Levenshtein distance and PER of the ``average" transcriptions obtained with our string averaging method for different subsets of the crowdsourced transcriptions. The ``average" transcriptions have higher quality than the original ones, especially when obtained from transcriptions of participants familiar with languages close to the target language. Providing alignments also improves the resulting ``average" transcription.}
    \label{tab:averages}
\end{table}

\section{Conclusion}

Through a small case-study, we show that crowdsourced transcriptions improve if the transcribers are provided with speech-to-translation alignment information, even if the alignments are noisy. Furthermore, we confirm the somewhat intuitive concept that workers familiar with languages closest to the language they are transcribing (at least phonologically) produce better transcriptions.

The combination of the mismatched transcriptions into one, using a simple string averaging method, yields even higher quality transcriptions, which could be used as training data for a speech recognition system for the endangered language. 
In the future, we plan to investigate the use of ROVER for obtaining a probabilistic transcription for the utterance, as well as explore ways to expand our phonologically aware string averaging method so as to produce probabilistic transcriptions, and compare the outputs of the two methods.

We plan to consolidate our findings by conducting case studies at a larger scale (collecting transcriptions through Amazon Turk) and for other language pairs. A larger collection of mismatched transcriptions would also enable us to build speech recognition systems and study how beneficial the improved transcriptions are for the speech recognition task. 

This work falls within an envisioned pipeline where we first align speech to translations, then crowdsource transcriptions, and last we train an ASR system for the endangered language. However, this is not the only approach we are considering. Another approach could replace crowdsourcing with multiple automatic phone recognizers (or even a ``universal" one) that would output candidate phonetic sequences, which we would then use to train an ASR system. Our main aim is to start a discussion about whether any additional information like the translations or the speech-to-translation alignments contain information that would help a human to interpret an endangered language, and how they could be used alongside the collected parallel speech for documentation efforts. 

We are also interested on how the annotation interfaces could be better designed, in order to facilitate faster and more accurate documentation of endangered languages. For example, our proposed interface, instead of just providing the alignments for each translation word, could also supply the transcriber with additional information, such as other utterance examples that this word has been aligned to. Or we could even attempt to suggest a candidate transcription, based on previous transcriptions that the transcriber (or others) have produced. This could potentially further improve the quality of the transcriptions, as providing several examples should improve consistency. Expanding our interface, so as to provide such additional information to the transcriber, is also part of our plans for future larger scale case studies.

\bibliographystyle{acl}
\bibliography{References}

\end{document}